# Author identification using multi-headed recurrent neural networks

Douglas Bagnall

douglas@halo.gen.nz

**Abstract.** A recurrent neural network (RNN) is used to model the language of several authors concurrently, with each author's text being represented by separate outputs that rely on a shared recurrent state. This allows the recurrent layer to model the language as a whole without over-fitting even with very small corpora. The method was developed for the PAN 2015 Author Identification task, where it performed well, coming first overall with an average AUC greater than .80.

## 1 Introduction

The PAN 2015 author identification task looks at the problem of deciding whether or not an document of unknown authorship was written by the author of small set of other documents. The known and unknown documents vary in topics or genres, and are spread across four languages—Dutch, English, Greek, and Spanish. A training corpus containing 100 problems in each language was provided. Software was tested using the TIRA[6] evaluation-as-a-service platform on an unseen evaluation corpus. Although the task is artificial, it is designed to be a proxy for common author verification problems.

### 1.1 Recurrent neural networks for language modelling

A recurrent neural network (RNN) is a feed forward neural network that shares parameters across time. At each time step $t$, a simple RNN has a hidden vector $h_t$ derived from the input vector $x_t$ and the previous hidden state $h_{t-1}$. The hidden vector is usually obtained via an affine transform offset by a bias vector $b_h$, followed by a non-linear "activation" function, $f_h$, which operates in a point-wise manner, giving the formula $h_t = f_h(W_{hh}h_{t-1} + W_{xh}x_t + b_h)$. The output vector $y_t$ is similarly obtained from the hidden state, with $y_t = f_y(W_{hy}h_t + b_y)$, though the non-linear function $f_y$ is often not point-wise. When a discrete probability distribution is desired, the "softmax" function

$$\sigma(\mathbf{z})_j = \frac{e^{z_j}}{\sum_k e^{z_k}}$$

is common because it gives a correct distribution (all values positive and summing to 1) and has useful properties when it comes to training the network. Figure 1 depicts a simple recurrent neural network.

The self-referential hidden state allows the network to model complex time series processes. Training it to do so involves iteratively adjusting the weights and biases, usually using some form of gradient descent and back-propagation through time (BPTT).

For the sake of brevity, the details of these algorithms are elided. Tomáš Mikolov's PhD thesis [4] offers a good introduction to these algorithms and the use of simple recurrent neural networks for language modelling. The language model predicts the flow of text, one symbol at a time, estimating a probability distribution for the $i$-th symbol $x_i$ given its predecessors, or $p(x_i|x_{i-1}, x_{i-2}, \ldots, x_1)$, where the symbols belong to a pre-determined vocabulary. In practice a language model has a limited horizon, basing its predictions based primarily on recent symbols. This is explicit in the case of n-gram based models, while recurrent neural networks can adapt the scope of their attention to suit the context.

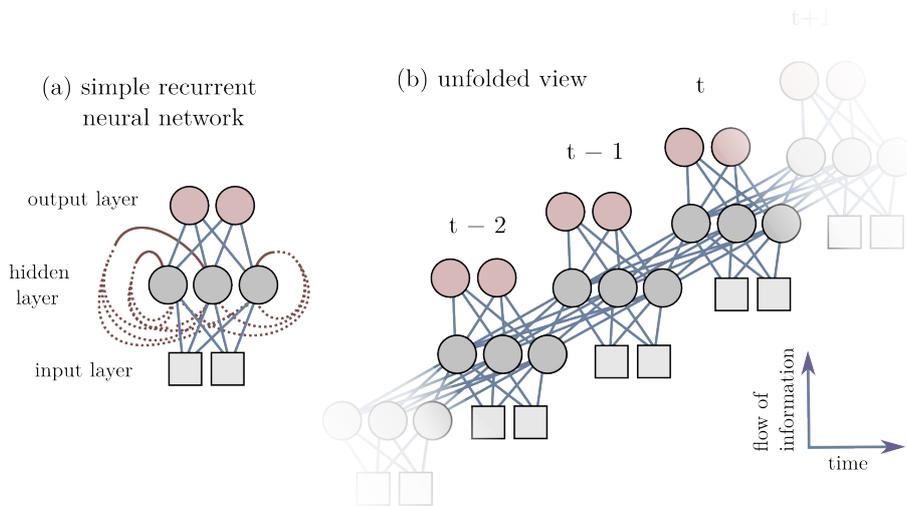

**Fig. 1.** The diagram on the left tries to show the output of the hidden nodes from the previous time step flowing round back into the hidden node inputs, but you are better off looking at the one on the right where the network has been unfolded through time. All layers are fully connected. At each time point the hidden state summarises the entire history of the input data.

Language models for are often word-based where the vocabulary is large but inevitably contains gaps. For alphabetic languages character-based models are also possible, and although they don't perform as well in general, they are robust in the face of novel words and require less text and time to train. Figure 2 shows a character-based language model trying to navigate its way through the character sequence `cat`.

The accuracy of a language model for a document can be measured as bits of cross entropy which is the mean negative binary log probability the model assigns to the symbols that actually occur. For a document of $N$ characters, this is $\frac{1}{N}\sum_i^N -\log_2 p(x_i|x_{i-1}, x_{i-2}, \ldots, x_1)$. Cross-entropy can be thought of as a measure of the information the model fails to model. A language model that predicts every character in a document with probability 1 will result in a zero cross-entropy; on the other hand assigning probability 0 to an occurring character is very costly.

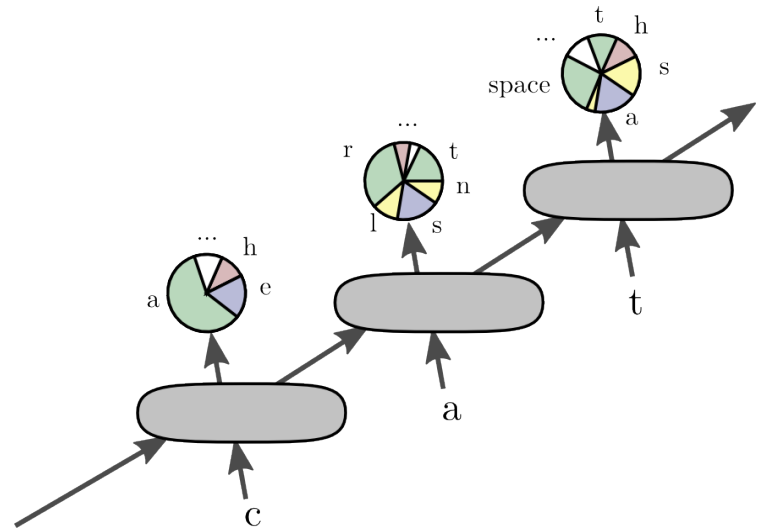

**Fig. 2.** An example of a character based language model at work. When the model sees a `c` it (taking into account previous unseen characters) predicts the an `a` as having about a 55% chance. The cross entropy of this single step is $-\log_2(0.55) \approx 0.86$. On seeing the `a` it then gives the `t` about a 20% chance, and $-log_2(0.2) \approx 2.32$. The cross entropy over these two steps is the mean, 1.59.

Supposing training is effective, a language model will better predict the flow of documents similar to the ones it was trained on—if that similarity is capturable by the model—and this will show up as reduced cross entropy relative to unrelated documents. The PAN 2015 author identification problem consists of several mini-corpora in various languages, [7][1] and each mini-corpus contains 1 to 5 documents known to be by a single author. These documents amount to a few thousand characters—generally fewer than in this paper—and the task is to decide whether another short document is by the same author.

The hypothesis underlying this work is that a character-level RNN language model trained on an author's known output will match that author's unknown output more closely than it matches text written by arbitrary others. Unfortunately conventional language models are trained on millions of characters, and would severely over-fit on a corpus of a few thousand—effectively learning to recite the training text verbatim. To combat this, a multi-headed character-level language model is introduced, which shares a common recurrent state but multiple independent softmax output groups. Each softmax group is trained predominantly on one author's corpus, causing the recurrent layer to model a combination of all the author's texts, approximating the language as a whole. The output groups learn to weight the recurrent values in a way that best reflects their authors' tendencies. Figure 3 attempts to illustrate this.

---

[1] See http://www.uni-weimar.de/medien/webis/events/pan-15/pan15-web/author-identification.html for more information on the PAN 2015 author identification task.

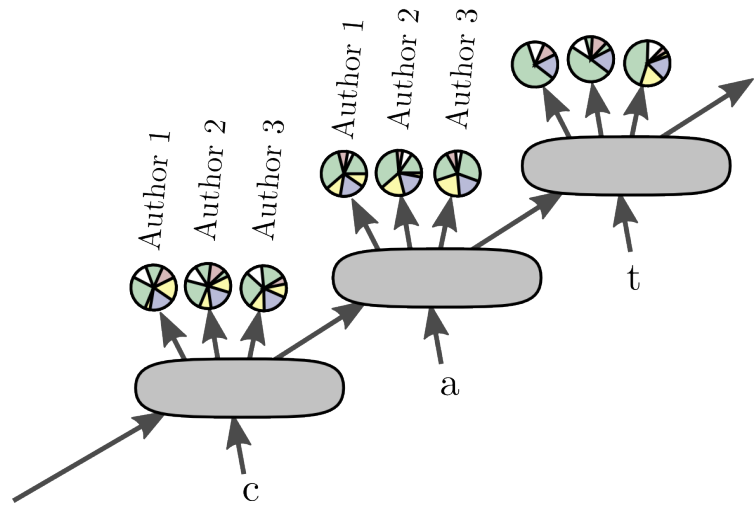

**Fig. 3.** The multi-headed RNN language model has multiple sets of output probabilities—one for each author—each making slightly different predictions. The cross entropies thus differ, with a relatively low cross entropy supporting the hypothesis that the corresponding author wrote the text being examined.

### 1.2 The PAN@CLEF 2015 author identification task

The relative cross-entropy scores are on their own insufficient to decide questions of authorship, but they can be brought as evidence to sway a pre-existing authorship question. The specific format of the PAN@CLEF 2015 author identification challenge shapes the interpretation of the scores in this paper. It should be understood that these methods are designed to perform well in a specific artificial task and may not be useful in general. It is therefore necessary to cover the task parameters that guide this work.

The task is split into four corpora, each based on a different language with varying levels of consistency in topic and genre, and with varying numbers and sizes of documents. Each corpus contains 100 or more questions, all of which have one or more documents by a "known" (but unidentified) author, and one document of "unknown" origin. The task is to decide whether each "unknown" document was written by the corresponding "known" author. Note that the task is symmetrical in cases with one "known" document — the same question is posed if the labels are swapped. It is known that exactly half the questions can be answered "yes" (that is, all documents are by the same author), and the other half "no". Looked at independently, each question has a "yes" probability of $0.5$, but if looked at together the probabilities are usefully correlated. There is a limited supply of each answer, so a confident "yes" in one question diminishes the chance of a "yes" in another. Any solution that ends up with an imbalance of positive and negative answers is guaranteed to be wrong.

Each answer takes the form of a score between zero and one, which is interpreted in a somewhat complicated fashion. The score for each language is the product of the area under the ROC curve (AUC), a general measure of binary classifier efficacy, and

C@1 which is a measure of accuracy which rewards admissions of uncertainty in some circumstances. Precisely, where $n$ is the number of problems, $n_c$ is the number of correct answers and $n_u$ is the number of explicitly uncertain answers:

$$c@1 = \frac{1}{n}(n_c + n_u n_c/n)$$

This formula reduces to plain accuracy $\frac{n_c}{n}$ if all answers are given definite answers (that is, $n_u = 0$). Marking a few answers as uncertain is better than arbitrarily guessing if accuracy is otherwise high.

For the purposes of C@1 in this context, a score greater than $0.5$ is regarded as "yes", a score less than $0.5$ is a "no", while exactly $0.5$ indicates uncertainty. Given the prior knowledge we have about the distribution of answers, a solution should never set more than half the scores on either side of $0.5$. The AUC measure is oblivious to score transforms that preserve monotonicity, so shifting the scores to suit the C@1 requirements while preserving AUC is possible if multiple uncertain answers are not required. A system that assigns scores from a uniform random distribution will tend to produce AUC and C@1 scores of $0.5$, hence an overall score of $0.25$. It is easy to devise a system that produces degenerate C@1 scores, but to do so with AUC requires a classifier that combines cleverness with perversity.

## 2 Method

### 2.1 Text preprocessing

The known and unknown texts are mapped to a smaller character set to reduce computational complexity and remove the overwhelming self-importance of extremely rare characters. A separate mapping is used for each language.

The text is first converted into the NFKD unicode normal form, which decomposes accented letters into the letter followed by the combining accent (for example, the code point `<U+00E0>` ("à") becomes `<U+0061><U+0300>` ("a" followed by a combining grave marker). Capital letters are further decomposed into an uppercase marker followed by the corresponding lowercase letter. Thus À becomes the triplet *<right-combining uppercase modifier>* a *<left-combining grave accent>*.

Various rare characters that seem largely equivalent are mapped together; for example the en-dash ("–") and em-dash ("—") are rare and appear to be used interchangeably in practice so these are mapped together. Other punctuation marks—such as the various forms of apostrophes and single quotation marks—are likewise collapsed into canonical forms. It is likely that this discards some usable information—indications of finger habits or choice of software—but the avoidance of extremely rare characters is important. If only one known author in the corpus uses a character, it may be assigned an excessive weight.

For the Greek text, all Latin characters are mapped to the letter s, arbitrarily chosen because it doesn't resemble a Greek character. The rationale is that foreign quotations and references appear too rarely for their content to be valuable and an attempt to model them would be wasteful, but the tendency to use them might be a useful signal. Following similar logic, all digits in all languages are mapped to 7. Runs of whitespace are collapsed into a single space.

At the end of this processing, any character with a frequency lower than 1 in 10,000 is discarded. Any characters occurring in a text but not in the resultant alphabet are ignored—there is no "unknown" token. Alphabet sizes are 39 for English, 47 for Dutch and Greek, and 51 for Spanish.

Finally, runs of more than 5 identical characters are truncated at 5. This is mainly aimed at the latin stretches in Greek text; there is little value in having the model try to guess the exact word length in a language it can't directly see. As an example, the following Greek text (the last paragraph of `GR094/known02.txt` in the training set):

```
«This», μουρμουρίζει χαμογελώντας μελαγχολικά ο Αμερικάνος,
ακούγοντας το Σαλονικιό «is the beginning of a beautiful friend-
ship».
Παίξτο ξανά.
```

maps to this:

```
«¹τsss», μουρμουρι²ζει χαμογελω²ντας μελαγχολικα² ο ¹αμερικα²νος,
ακου²γοντας το ¹σαλονικιο² «ss sss sssss ss s sssss sssss».
¹παι²ξτο ξανα².
```

where the superscript ¹ represents a capital marker attaching to the next character, while the superscript ² is an acute belonging to the previous character. In this case the transformation reveals that the author has spelt "This" with an uppercase τ (tau) instead of the visually identical uppercase t.

The character mappings were settled before training started and no attempts were made to test their efficacy.

### 2.2 Multi-headed recurrent neural network language model

This concept was briefly introduced in section 1. A single recurrent neural network is trained to predict the flow of text by many authors while sharing a collective model of the complete language. The output layer has a softmax group that models the probability distribution expected for each author, and sometimes another for a "control corpus"—a large body of text intended to help prevent over-fitting in the recurrent layer. For convenience the PAN 2014 training set is used for the control corpora. It turns out that there is some overlap between the 2014 and 2015 training sets (and possibly test sets), so the control corpora are not completely independent. Empirically they seem to have a very small positive effect.

### 2.3 ReSQRT activation

The activation function used for the recurrent layer is a rectified shifted square root, or ReSQRT, defined as

$$f(x) = \begin{cases} \sqrt{x+1} - 1 & \text{if } x \geq 0 \\ 0 & \text{otherwise.} \end{cases}$$

The derivative is $\frac{1}{2\sqrt{x+1}}$ for $x > 0$, and otherwise 0. In terms of $y = f(x)$ (which is of practical use in training) the non-zero part is $\frac{1}{2(y+1)}$. This follows the model of the widely used rectified linear unit (ReLU,[5] defined as $f(x) = max(x, 0)$) in that the output and derivative is zero for all non-positive numbers, which offers performance and propagative benefits and allows neurons to opt out of opining on areas outside their speciality. ReLU can be difficult to use in recurrent neural networks as the lack of inherent scale means it can slip into an explosive cycle of amplification more easily than conventional squashing activations like $tanh$. ReSQRT steers a middle course, and empirically works well for character based language models. The ReSQRT activation may not have been described before.

The training uses a form of adagrad learning, where the rate at which a weight can change is inversely proportional to the $L^2$ distance that it has already changed. This amounts to a monotonically decreasing per-weight learning rate.

### 2.4 Training

Training is structured as a number of "sub-epochs". In a sub-epoch each author provides a training text. Where there is one text for every author, a sub-epoch is the same as a conventional epoch; otherwise the author's texts are used in a cyclical way. In some runs all the texts of each author are concatenated—in these cases the sub-epoch is also a true epoch. In another mode, each sub-epoch is "balanced", with documents being drawn from each author until each has been trained around the same amount.

At each training step there is a chance of the training example "leaking" and affecting other authors, as if they had also made that particular choice of character. That is, calling the $N$ authors of the $N$ problems $a_1, a_2, \ldots, a_n$, if a character $c$ is from a text by $a_i$, in the normal case the back-propagated error for all other authors $a_{j \neq i}$ is zero; when an example $j$ leaks the training error for $a_j$ is back-propagated. This has the effect of blaming each author not only their actual text, but in small part for that of all of the others, which serves as a proxy for the overall possibilities of the language. This avoids extreme overfitting. The initial leakage rate is in the order of $1/N$, and it decays exponentially with each sub-epoch. Towards the end of training, the leakage rate is very low and each author's sub-model is being trained almost entirely on its own text. The parameters are thus roughly set during early training with high leakage and high learning rate, then refined to specialise on the author's style.

A mini-batch size of 40 is used, meaning the weights are modified by the accumulated deltas after every 40 characters. The training gradient for first 10 characters of each text is ignored. Back-propagation through time (BPTT) is truncated at a depth of 70, or sooner when the gradient diminishes below an adaptive threshold. The hidden layer is quite small; experiments with larger sizes showed no benefit.

Where there is a single "known" text, and it is significantly shorter than the "unknown" text, the two are swapped around and the RNN is trained on the unknown text. Insufficient training text, which causes over-fitting, is worse than insufficient test text which only increases uncertainty.

The recurrent neural network code is written in C while the PAN specific parts are Python.[2]

### 2.5 Ensemble of variations

The final results combine several runs together, using a variety of configurations of meta-parameters. Approximate ranges are shown in Table 1. Each run consists of a training phase lasting several minutes or hours followed by the calculation of cross entropies for the unknown documents, which takes less than a second.

| meta-parameter | typical values |
| --- | --- |
| initial adagrad learning scale | 0.1, 0.14, 0.2, 0.3 |
| initial leakage between classes | $\frac{1}{4N}$ to $\frac{5}{N}$ |
| leakage decay (per sub-epoch) | 0.67 to 0.9 |
| hidden neurons | 79, 99, 119, 139 |
| presynaptic noise $\sigma$ | 0, 0.1, 0.2, 0.3, 0.5 |
| sub-epochs | 6 to 36 |
| text direction | forward or backward |
| text handling | sequential, concatenated, balanced |
| initialisation | gaussian, zero |
| control corpus | PAN14 or none |

**Table 1.** Meta-parameters used in training. Due the use of randomness, and the hidden nature of the evaluation, it is not precisely known which combinations of meta-parameters were used in the final ensembles.

Sets of seemingly reasonable meta-parameters are chosen per-language via a haphazard search, and random selections of these configurations are used in the the evaluation ensembles. [3] The evaluations are run with a time-out; after a set number of hours all the test scores obtained are averaged together. The length of time dedicated to each ensemble was determined by the available time before the submission was due.

In tests with the training set (and in line with received wisdom), ensembles performed slightly better than the majority of their constituents.

### 2.6 Interpretation

With $N$ predictive output clusters representing the $N$ "known" authors in the problem corpus, the system produces $N$ cross entropy scores for each document. For the training data there are 100 problems, hence 100 of these scores, for each language; in the evaluation set the numbers vary between languages. The problems may share known authors so in reality the number of authors may be smaller. The scores are not directly comparable to each other as there is both random variation in the various sub-models' performance and inherent variation in the texts' predictability. Some texts are more difficult for all models. The scores are normalised by subtracting the mean cross-entropy for each text, and the resultant zero-centred relative entropies are scaled so that each sub-model has scores in a similar range. For each author's sub-model, the ranking of the score for the unknown text determines the probability that the author wrote the text. For example, if a text ranks first (i.e. has the lowest score) it is likely to be that author's work; if it ranks last, it is probably not. This system is likely not optimal, given it ignores nuances in the evidence, but it is simple to implement and works well enough.

---

[2] See https://github.com/douglasbagnall/recur and https://github.com/pan-webis-de/caravel for software details.

[3] The configuration pools from which the ensembles were chosen can be found at https://github.com/pan-webis-de/caravel/tree/master/config. Some decoding is necessary.

The final score must take the form of a probability between 0 and 1, with 0.5 having special significance for the C@1 calculation. As discussed in section 1.2, the task is designed so that in exactly half the cases the unknown documents are by the known author, so (supposing confidence in the model) the correct way to align the 0.5 point is to take the median. Documents ranking better than the median receive a score above 0.5 linearly related to their ranking, while those ranking worse than the median receive a lower score. For the Spanish, Greek, and English tasks a small radius around the median was also assigned an 0.5 score. With the training corpus this had little effect on the AUC and slightly increased the C@1 component.

## 3 Results

Results are shown in Table 2 and Table 3.

| Team | Dutch | English | Greek | Spanish | micro | macro |
| --- | --- | --- | --- | --- | --- | --- |
| **Bagnall** (this paper) | 0.451 | **0.614** | **0.750** | 0.721 | **0.608** | **0.628** |
| Moreau et al. | **0.635** | 0.453 | 0.693 | 0.661 | 0.534 | 0.606 |
| Pacheco et al. | 0.624 | 0.438 | 0.517 | 0.663 | 0.480 | 0.558 |
| Hürlimann et al. | 0.616 | 0.412 | 0.599 | 0.539 | 0.487 | 0.538 |
| Bartoli et al. | 0.518 | 0.323 | 0.458 | **0.773** | 0.417 | 0.506 |

**Table 2.** Selected results from the 2015 PAN Author Identification evaluation, taken from [7]. The score is the product of the area under the ROC curve (AUC) and the C@1 score using 0.5 to indicate a non-decision, as described in section 1.2. The results are sorted by macro-average. Micro-average differs from macro-average because there are different numbers of problems in each languages: 500 for English, 165 for Dutch, and 100 each for Spanish and Greek.

Relative to the scores obtained in experiments with the training set, the evaluation score for Greek is higher, while that for Spanish is lower. The English and Dutch results are in the expected range.

| language | AUC | C@1 | score | margin | rank | runtime |
|---|---|---|---|---|---|---|
| Dutch | 0.70 | 0.64 | 0.45 | -0.18 | 7 | 12:00:43 |
| English | 0.81 | 0.76 | 0.61 | 0.09 | 1 | 21:44:03 |
| Greek | 0.88 | 0.85 | 0.75 | 0.06 | 1 | 10:07:49 |
| Spanish | 0.89 | 0.81 | 0.72 | -0.05 | 2 | 11:21:41 |

**Table 3.** Details of this paper's results, showing the constituent AUC and C@1 portions, and the margins relative to the highest scores from other teams. A positive margin indicates this software's score was ahead of the nearest team; when negative the margin shows how far it was behind the leader.

## 4 Discussion

The consistently bad results for the Dutch task seem to reflect a drastic genre difference between the known and unknown texts. Many teams did a lot better in this task, presumably by not concentrating on the local flow of characters and instead using linguistic knowledge or otherwise deriving longer range features.

On the other hand, where the genre and topic seem closer, this model performs very well despite using no specialist linguistic or natural language processing techniques. It ought to extrapolate well to other languages, and work well in concert with unrelated methods.

The performance in the English task is notable, exceeding other teams' results by a significant margin. Possibly this is due the larger corpus (500 problems) increasing the breadth of training and perhaps smoothing out noise in the ranking procedure.[4]

Without an active recurrent layer the model falls to predicting unigram character probabilities per author via the output bias vector. The cross entropy in this case amounts to weighting the frequencies of characters in the text (using what amounts to a form of logistic regression). When (accidentally) run this way on the training set, the AUC score was $0.85$ for English and $0.91$ for Spanish; for English this was one of the best training results. While this is humiliating for the RNN, it confirms the validity of the training and scoring methods. Leakage between authors allows each sub-model to learn background probabilities for the language as a whole, which is presumably better than the common technique of giving unseen symbols a fixed low chance. By being entirely additive, cross entropy sidesteps the curse of dimensionality. It also eschews (perhaps too much) the possibility of an *a ha!* moment involving strong positive evidence—positive evidence is typically accumulated a fraction of a bit at a time over the course of the entire document. In contrasts a typical human approach often involves identifying idiosyncratic usages and ignoring the boring bits in between (perhaps an RNN would do well in ensembles with people). An RNN that directly reported an authorship probability distribution is conceivable but is unlikely to be easy to train. The character level language model learns to concentrate on the text; a more direct approach is likely to be more distracted. Tirelessly trying to understand the text is the strength of this method.

---

[4] PAN competitors have no access to the test data or detailed results. These comments rationalising varying levels of success are purely speculative.

Most contemporary research into recurrent neural networks focuses on more sophisticated models using (for example) long short term memory (LSTM [2]) and gated recurrent units (GRU [1]), with far deeper and wider architectures than seen here. These allow RNNs to discover longer range and more complex dependencies. Whether that would help in this task is an open question—a long range view is not much use when training on a 300 character text. The ReSQRT activation used here may work well precisely because it does not try to learn too much, and instead relies on the surface detail of writing style. The computational cost of more sophisticated networks would also be problematic in the context of a PAN competition with limited hardware.

Nevertheless, this work demonstrates that even small recurrent neural networks can be a useful tool in authorship identification. The approach owes more to information theory than traditional clustering, and happily avoids the quagmires of feature selection and excessive dimensionality. Results with the training set, where the size of the recurrent layer was mistakenly reduced to zero, suggest that the interpretive approach is more important than the predictive ability of the network. For that I thank the recently deceased David J. C. MacKay [3], whose clear writing on topics related to inference guided this work and much else.